# Knowledge distillation as a pathway toward next-generation intelligent ecohydrological modeling systems

Long Jiang[1,*], Yang Yang[2,3], Ting Fong May Chui[2], Morgan Thornwell[4], Hoshin Vijai Gupta[5]

[1]*Puget Sound Institute, University of Washington, Tacoma, Washington, USA*

[2]*Department of Civil Engineering, The University of Hong Kong, Hong Kong SAR, China*

[3]*School for the Environment, University of Massachusetts Boston, Boston, Massachusetts, USA*

[4]*Thornwell Labs, Portland, Oregon, USA*

[5]*Department of Hydrology and Atmospheric Sciences, The University of Arizona, Tucson, Arizona, USA*

*Email: jlon@uw.edu*

## Keywords

Cognitive modeling; Graph neural network (GNN); Process-based model; Machine learning; Residual learning; Transfer learning

## Abstract

Simulating ecohydrological processes is essential for understanding complex environmental systems and guiding sustainable management amid accelerating climate change and human pressures. Process-based models provide physical realism but can suffer from structural rigidity, high computational costs, and complex calibration, while machine learning (ML) methods are efficient and flexible yet often lack interpretability and transferability. We propose a unified three-phase framework that integrates process-based models with ML and progressively embeds them into artificial intelligence (AI) through knowledge distillation. Phase I, behavioral distillation, enhances process models via surrogate learning and model simplification to capture key dynamics at lower computational cost. Phase II, structural distillation, reformulates process equations as modular components within a graph neural network (GNN), enabling multiscale representation and seamless integration with ML models. Phase III, cognitive distillation, embeds expert reasoning and adaptive decision-making into intelligent modeling agents using the Eyes-Brain-Hands-Mouth architecture. Demonstrations for the Samish watershed highlight the framework's applicability to ecohydrological modeling, showing that it can reproduce process-based model outputs, improve predictive accuracy, and support scenario-based decision-making. The framework offers a scalable and transferable pathway toward next-generation intelligent ecohydrological modeling systems, with the potential extension to other process-based domains.

## 1 Introduction

Ecohydrological modeling, broadly defined to encompass ecological and hydrological processes and their couplings, is essential to understanding complex environmental systems dynamics. It further enables scenario analysis to evaluate the potential impacts of land-use change, extreme events, and management interventions under accelerating climate change and human pressures *(Bonetti et al., 2021; Porporato et al., 2015)*. Process-based ecohydrological models have long been central tools for streamflow forecasting, nutrient transport analysis, and ecosystem management, owing to their physical interpretability and process fidelity *(Clark et al., 2017; Fatichi et al., 2016)*. However, their development and application remain constrained by structural rigidity, high computational costs, and the need for expert, case-specific calibration *(Beven, 2018; Mizukami et al., 2019; Nepal et al., 2017)*. which can restrict their scalability and broader adoption *(Addor & Melsen, 2019; Garzón et al., 2022; Hutton et al., 2016)*.

These limitations have driven growing interest in data-driven methods, particularly machine learning (ML), which offers flexibility and computational efficiency in predicting streamflow, nutrient loads, and ecological responses, as well as in parameter inversion *(Kratzert et al., 2019; Reichstein et al., 2019; Shen et al., 2021)*. However, purely ML models often overlook physical laws, spatial coupling, and causal dependencies, limiting interpretability and credibility *(Nearing et al., 2021)*. In high-stakes applications, process-based models remain indispensable for physical consistency *(Wiggerthale & Reich, 2024)*. At the same time, the scarcity of long-term, high-quality data, especially for water quality and ecological variables, further constrains the transferability of ML models across basins *(Klotz et al., 2022; Kratzert et al., 2023; Zhi et al., 2024)*.

A deeper challenge lies in the limited conceptual adaptability of both paradigms to new processes and scenarios. Process models are costly to modify due to structural rigidity *(Clark et al., 2017)*, often constrained to limited model libraries or requiring fundamental changes to source code. ML models often fail in unseen scenarios where relevant processes or conditions are absent from the training data, necessitating retraining with additional data that may not be readily available *(Banda et al., 2022)*.

Thus, neither paradigm alone is sufficient, making it essential to combine their strengths by extracting knowledge from traditional process-based models to enhance ML capabilities *(Massoud et al., 2023)*. Recent efforts have integrated such domain knowledge into ML models, for example by applying physical constraints to regularize learning and improve extrapolation *(Feldman et al., 2023; Raissi et al., 2019; Reichstein et al., 2019)*. Advances such as physics-informed neural networks and theory-guided data science have demonstrated that incorporating mechanistic constraints can enhance the efficiency, robustness, and scientific credibility of ML models *(Karpatne et al., 2017; Read et al., 2019; N. Wang et al., 2020; Y.-H. Wang & Gupta, 2024a, 2024b)*. However, such constraints (e.g., water or energy balance) cover only a small fraction of the knowledge embedded in process models. Broader ecohydrological knowledge encompasses not only explicit process equations and empirical parameters, but also cognitive frameworks distilled from decades of expert practice, including heuristic methods, reasoning patterns, modeling strategies, and decision rules *(Clement, 2022; Dedhia et al., 2025).* These implicit forms of expertise, together with explicit mechanisms, constitute the foundation for understanding and simulating ecohydrological systems. Yet due to technical barriers, much of this valuable knowledge remains underutilized and is rarely transformed into reusable modeling resources.

Against the backdrop of rapid advances in large language models and artificial intelligence (AI), future ecohydrological models should move beyond merely fitting observational data to incorporate cognitive strategies, automate simulation workflows, and provide transparent explanations of their predictions. Achieving this goal requires integrating explicit process-model knowledge with implicit expert cognitive frameworks within ML architectures to guide modeling and support informed decision-making *(Eythorsson & Clark, 2025; Knoben & Spieler, 2022; Nearing et al., 2021)*. However, mature methodologies and systematic studies for extracting and embedding such knowledge into modular, transferable AI architectures remain scarce *(Maity et al., 2024)*.

Building on this momentum, we propose a three-phase knowledge distillation framework that systematically transfers process-based ecohydrological understanding into AI architectures (**Fig. 1**). It distills knowledge across three levels: behavioral patterns from existing models, structural representations of ecohydrological processes, and cognitive frameworks for model configuration and interpretation. While ultimately aimed at autonomous, expert-emulating AI systems for ecohydrological modeling, the framework enables a gradual and interpretable transition for traditional modelers, allowing them to harness the efficiency and adaptability of AI without discarding decades of accumulated knowledge. To our knowledge, systematic, multi-level distillation of this kind has not yet been reported in ecohydrology.

*Phase I: Behavioral Distillation*

The first phase is designed as a pragmatic entry point for process-based modelers to immediately benefit from ML's computational advantage, helping to overcome the challenges of high-dimensional parameter spaces and intensive model runtimes. Unlike direct surrogate modeling, behavioral distillation does not seek to fully replace process-based models. Instead, it leverages ML to extract key input–output behavioral patterns from complex ecohydrological systems, serving as a low-barrier yet effective enhancement strategy for traditional models.

To achieve this, we introduce a two-step strategy that combines model simplification with ML. First, a simplified variant of the process-based model is constructed to preserve essential process fidelity while efficiently generating training data for the behavioral-distillation surrogate. Second, the surrogate is then fine-tuned using transfer learning (or residual learning) to improve predictive accuracy and better replicate the behavior of original simulations. This two-step approach enables rapid simulation, parameter estimation, and scenario exploration, while retaining a process-based, physically consistent foundation.

Surrogate models (e.g., LSTM, MLP, CNN) have been widely used to approximate the input–output behavior of process-based models across diverse conditions. They deliver rapid simulation at lower computational cost and have achieved considerable maturity in applications such as climate-driven runoff prediction *(Mohammadi, 2021; Xiang et al., 2020)*. However, most existing efforts focus on direct emulation. Systematic behavioral distillation remains underexplored and is expected to become a major research focus in the next 3–5 years.

*Phase II: Structural Distillation*

Structural distillation refers to the abstraction of process-based ecohydrological models into flexible, trainable ML architectures. It addresses limited interoperability across models, stemming from the absence of standardized representations of ecohydrological processes. This phase extracts and modularizes core equations and data flows from established models to form unified, interoperable process components agnostic to programming language, computational platform, and model architecture.

To this end, we propose a graph-based modeling framework grounded in core ecohydrological processes, released as an open-source PyTorch implementation, *EcoHydroModel*. This framework employs graph neural network (GNN) to achieve structural abstraction by decomposing complex process-based models into reusable, node-level process equation modules. It preserves core physical relationships while enhancing adaptability to diverse data conditions, enabling knowledge transfer across systems, regions, and spatial resolutions. As an emerging direction, structural distillation is expected to evolve into a mature and robust methodology within the next five years.

*Phase III: Cognitive Distillation*

Cognitive distillation refers to the systematic embedding and integration of expert knowledge (including heuristic methods, reasoning patterns, modeling strategies, and decision rules) into AI systems. It represents a paradigm shift in ecohydrological modeling from passive behavioral emulation to autonomous modeling infused with domain expertise.

Achieving this goal necessitates the explicit formalization of modeling assumptions, structural choices, and reasoning logic that are often implicit and context dependent in expert practice. This is an ambitious objective that brings substantial theoretical and technical challenges. In this paper, we offer a preliminary exploration of the technical pathways and methodological architectures needed, while identifying key challenges and recommending potential solutions.

Although still conceptual, this phase aims to develop AI agents capable of autonomously constructing, dynamically adapting, and continuously optimizing ecohydrological models, thereby laying the foundation for knowledge-driven, self-evolving intelligent modeling systems. We anticipate mature applications of this approach to emerge within the next decade.

*Practical Demonstration*

To illustrate the framework's applicability and workflow, we present case demonstrations in the Samish watershed (Puget Sound, Washington, USA). The 220 $km^2$ watershed exhibits diverse hydrology and land use, pronounced spatiotemporal coupling of processes, and long-term observations, making it well suited for this demonstration.

*Outline & Scope of the Paper*

Section 2 presents Phase I (behavioral distillation), using ML to distill key behaviors from process-based models. Section 3 presents Phase II (structural distillation), which abstracts and unifies model structures into modular components. Section 4 explores Phase III (cognitive distillation), embedding expert knowledge within adaptive modeling agents. Section 5 discusses the issues of overfitting and overparameterization, and examines broader practical implications and potential extensions. Finally, Section 6 synthesizes the contributions to ecohydrological modeling.

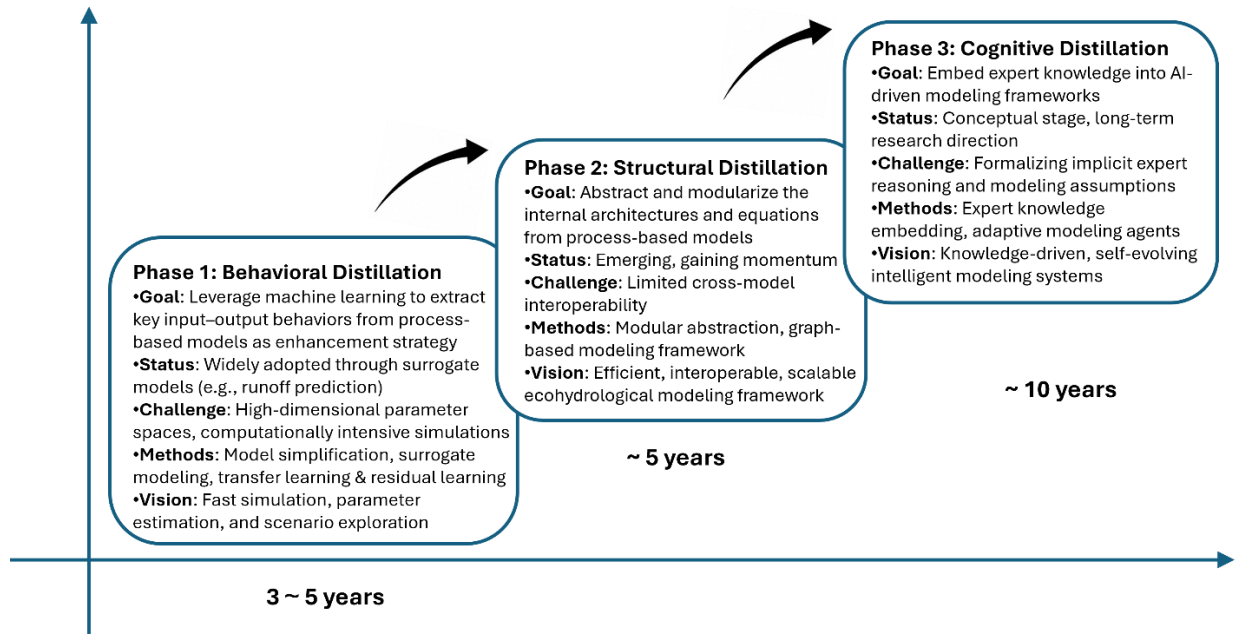

*Fig. 1 Three-phase roadmap for knowledge distillation from ecohydrological modeling*

## 2    Phase I: Behavioral Distillation

### 2.1    Toward Behavioral Distillation: A Pragmatic Entry Point for Model Enhancement

Traditional process-based ecohydrological models together with their well-established, long-running instances have been extensively tested and are widely regarded as reliable tools. Certain stable, repeatedly validated parameters have become core knowledge assets of these models. However, two persistent challenges remain: escalating computational cost with finer spatial–temporal resolution, and increasing calibration complexity as parameter dimensionality grows.

Given ML's strengths in pattern recognition and function approximation, we propose ML as a practical entry point for traditional modelers to immediately harness computational advantages and address these challenges. To this end, we introduce the "*behavioral distillation*" framework, which combines model simplification with ML, providing a low-barrier yet effective strategy to enhance traditional models. Unlike direct surrogate modeling, behavioral distillation does not seek to fully replace the process model. Instead, it trains ML models on simplified model inputs/outputs and truncated simulation trajectories generated by process-based models to reproduce the key behavioral patterns of complex systems. This approach significantly reduces the computational cost of generating training and evaluation data while preserving physical realism. It also enables immediate practical gains from ML-augmented modeling in targeted applications, without requiring spatial transferability at this initial stage.

As shown in **Fig. 2**, the behavioral distillation framework comprises pathways based on model simplification: (i) direct surrogate modeling, (ii) residual learning, (iii) transfer learning, and (iv) a hybrid strategy. Model simplification proceeds either via (a) *resolution coarsening*, which replaces the original high-resolution setup with a coarser spatial or temporal grid, or (b) *process reduction*, which either simplifies complex processes within the model or substitutes them with functionally similar but less complex models. It is aimed at transforming the original complex models into more computationally efficient variants that preserve essential process fidelity. These simplified models can efficiently generate large amounts of representative low-precision or short-term simulations at lower computational cost, which can then be used to train ML models that capture core system dynamics.

Building on these simplified models, residual learning, transfer learning, and hybrid strategies can be developed, each with distinct advantages in mechanism, efficiency, and applicability.

- Residual learning targets the systematic differences between simplified and original model outputs by learning a corrective function for residual errors, rather than reconstructing a full surrogate. However, it requires large paired original–simplified datasets, making it more suitable for medium-complexity cases when the original model can be run economically.

- Transfer learning follows a pretrain–finetune scheme: pretrain a surrogate on simplified-model data, then adapt it with limited original model outputs. In early stages, it relies mainly on low-cost data from the simplified model, with minimal dependence on original model outputs, which suits expensive models that cannot be run at scale. Its effectiveness, however, depends on the dynamic similarity between simplified and original models, and performance declines when differences are large.

- The hybrid strategy integrates residual and transfer learning: residual learning corrects major systematic biases, while transfer learning aligns the surrogate with the original domain. This combination lowers dependence on original model outputs and improves robustness, making it well suited to complex, high-cost tasks that demand both accuracy and efficiency.

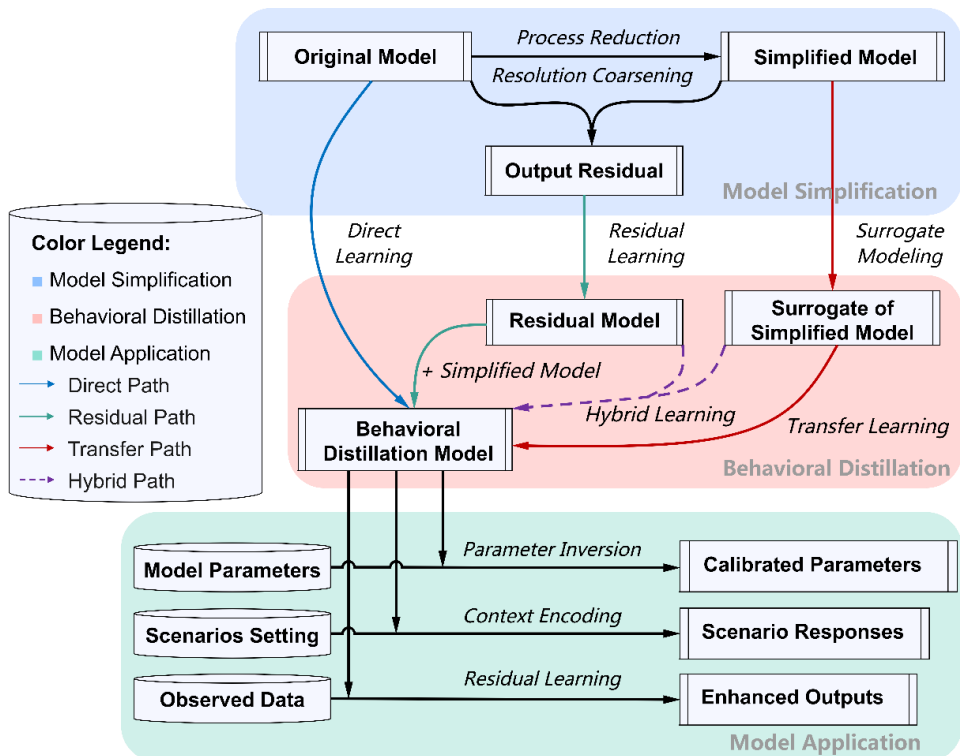

*Fig. 2 Behavioral distillation via model simplification and machine learning*

## 2.2    Applications for Behavioral Distillation Models

With additional parameter information or training data, distilled models can also support parameter estimation, scenario analysis, and output enhancement.

For parameter estimation, traditional calibration often relies on predefined or literature-based parameter values, manual tuning based on expert experience, or global optimization techniques such as simulated annealing and genetic algorithms. The latter typically require numerous costly model runs. Incorporating contextual information relevant to the spatiotemporal variability of model parameters into the training inputs can enable the distilled model to explicitly learn the mapping between parameter configurations and system behavior. This enables rapid calibration through global optimization or differentiable parameter inversion, substantially reducing computational costs.

For scenario analysis, including drivers such as climate indicators, land-use change, or extreme events in the inputs can allow the model to learn system behavioral responses to changes in external forcings. This supports high-fidelity scenario simulations without re-running the original model, improving adaptability for large-scale policy or management evaluations.

For output enhancement, a secondary residual learning layer can be introduced to align distilled model outputs with observations, correcting biases from structural simplifications, parameter uncertainties, or unmodeled external disturbances. This improves realism without altering the base model and offers interpretability for diagnosing deficiencies, which is especially valuable when high-quality observational data are scarce.

## 2.3    Simple Case Demonstration

Using the Samish watershed (**Fig. 3(a)**) as a case study, we applied ML for behavioral distillation of ecohydrological processes, focusing on streamflow (mm/day) and nitrate nitrogen loss (g N/m$^2$/day). Two representative models were used: the semi-distributed Soil and Water Assessment Tool (SWAT) and the fully distributed Visualizing Ecosystem Land Management Assessments (VELMA). Two approaches for model simplification were applied. In the resolution coarsening, the number of SWAT hydrologic response units (HRUs) was reduced from 1,896 to 187, and the VELMA spatial resolution was coarsened from 30 m to 360 m. In the process reduction, the hydrological module was replaced with the lumped Hydrologiska Byråns Vattenbalansavdelning (HBV) model, and a simplified nitrogen cycle module was added for water quality simulation. Both transfer learning and residual learning were implemented using multi-layered LSTM models with identical architectures. Detailed watershed characteristics and model configurations are provided in the supplementary materials.

**Figs. 3(b)–(i)** compare the original models with different behavioral distillation models, evaluated by predictive accuracy and computational efficiency. Accuracy was evaluated using a composite score, defined as the average of Nash–Sutcliffe efficiency (NSE) and Kling–Gupta efficiency (KGE). Computational efficiency was measured by running time, including model training and prediction.

- **Overall accuracy:** All models reproduced runoff with reasonable accuracy. Even direct surrogate models maintained high predictive skill (**Figs. 3(f)–(i)**). Differences were more evident for nitrate nitrogen loss, particularly during peak events (**Figs. 3(j)–(m)**).
- **Impact of model complexity:** For the highly complex VELMA model, direct surrogate and residual learning models trained solely on original outputs achieved poor generalization (scores below 0.6). In contrast, the SWAT-based models reached scores close to 0.9. This difference is largely due to SWAT's lower spatial heterogeneity and fewer HRUs, which simplify input–output relationships and reduce the learning burden for ML models.
- **Strategy comparison:** Residual learning and transfer learning performed similarly overall. For complex process-based models like VELMA in highly heterogeneous regions, transfer learning with resolution coarsening was more stable and effective than residual learning with process reduction. High process complexity can cause process reduction to distort model responses in ways that are difficult to recover, whereas resolution coarsening tends to preserve the main processes. Moreover, high heterogeneity can introduce residual noise unrelated to input drivers, complicating residual training.

- **Computational efficiency:** Direct surrogate models required the least runtime, followed by process-reduction-based residual learning. For resolution-coarsening approaches, most computational cost came from data generation for training rather than ML prediction. Although involving more steps, the hybrid strategy combining process reduction with residual learning reduced total computational costs substantially.

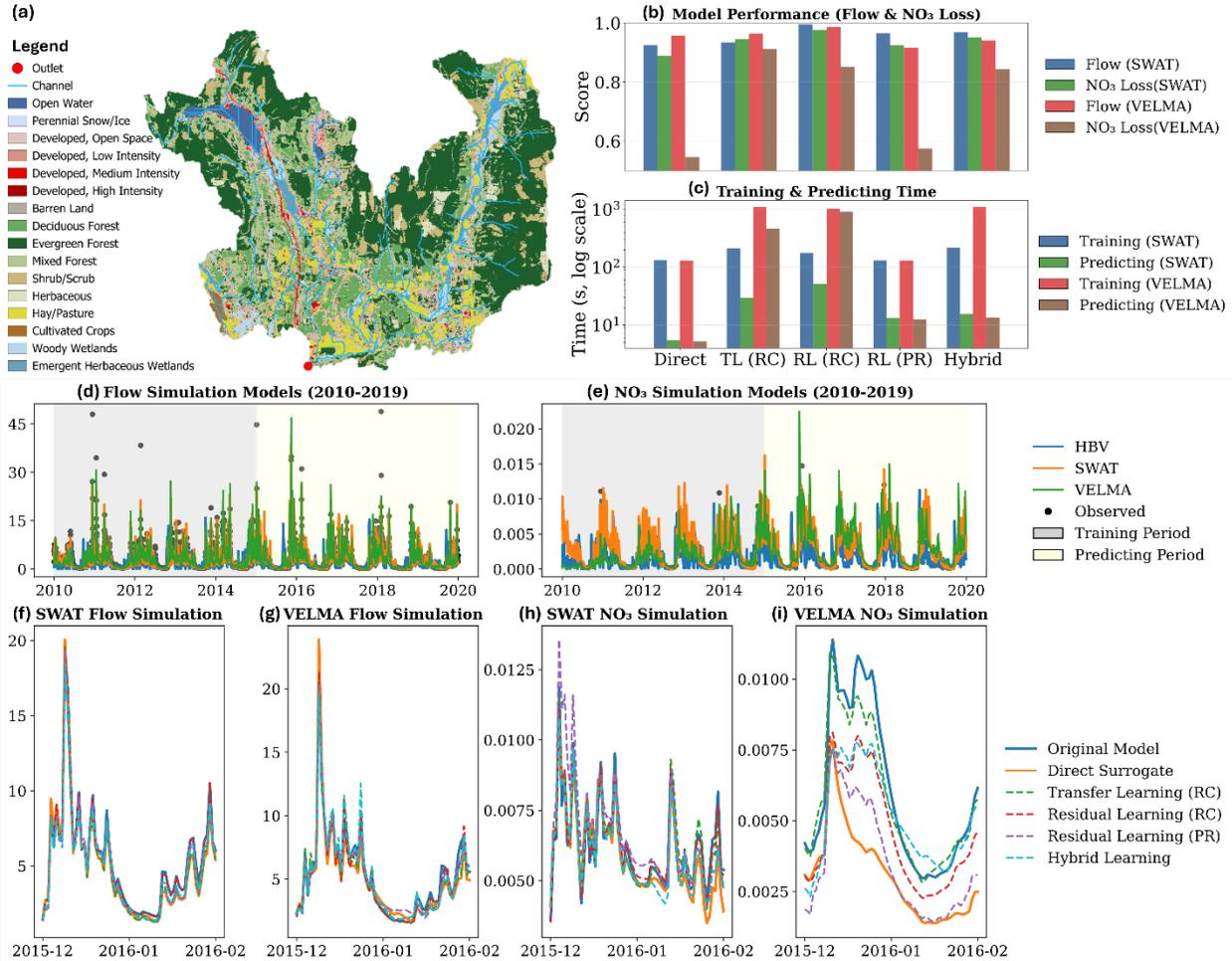

*Fig. 3 Comparison of predictive accuracy and computational efficiency for the original models and different behavioral distillation strategies. RC: resolution coarsening, PR: process reduction, TL: transfer learning, RL: residual learning. (a) Land use distribution and outlet in the Samish subbasin. (b–c) Model performance comparisons using the same x-axis labels; (b) Composite scores (average of NSE and KGE) for streamflow (mm/day) and nitrate nitrogen loss (gN/m²/day). (c) Total computational time for training and predicting (log scale). (d–e) Long-term simulations from the original process-based models. (f–g) Zoomed-in predictions of streamflow from behavioral distillation models. (h–i) Zoomed-in predictions of nitrate–nitrogen loss from behavioral distillation models.*

**Key conclusions:**

1. For relatively simple processes such as streamflow in this case and soil moisture *(Aieb et al., 2025; Sun et al., 2023)*, direct surrogate models can capture long-term dynamics with high accuracy. By contrast, for variables with high process complexity and strongly affected by spatial heterogeneity (e.g., nitrate nitrogen loss), more specialized behavioral distillation strategies are needed.

2. Residual learning and transfer learning based on model simplification performed similarly in this case. Transfer learning was more effective for the resolution coarsening scenarios, where model structures remained the same or similar but spatial heterogeneity was high. Residual learning seems better suited for process reduction scenarios that introduce structural differences.

3. Hybrid strategies that combine residual correction and transfer adaptation on top of simplified models can maintain high accuracy while substantially reducing computational cost, offering a practical accuracy–efficiency trade-off for complex systems.

Overall, applying behavioral distillation as an enhancement to the process-based models in the Samish watershed improved computational efficiency while preserving high-fidelity simulation of key ecohydrological processes (e.g., runoff generation and nutrient transport). Building on this approach enables rapid gradient-based parameter calibration and accelerated scenario analysis. Although the models here were tailored to the Samish watershed and cannot be assumed to generalize directly, the proposed framework is highly transferable and, with region-specific adjustments, can be applied to large-scale watershed management and decision support.

## 3 Phase II: Structural Distillation

### 3.1 Toward Structural Distillation: Learning Directly from Process Equations

Although Phase I behavioral distillation can significantly improve the practical utility and computational efficiency of process-based models through ML, its generalization capacity can be limited. First, such models often encode region-specific response patterns into their parameters, reducing accuracy when transferred elsewhere. Second, process-based models often depend on structural simplifications of climate, soils, and land use (e.g., a four-layer soil column in VELMA). These inherent constraints limit the effectiveness of behavioral distillation, leading to performance declines when transferred to regions or scenarios where the underlying assumptions are no longer appropriate.

Moreover, the ecohydrological modeling frameworks remain highly fragmented. Models differ in structural assumptions, computational platforms, programming languages, and process representations *(Brewer et al., 2018)*. These differences increase the barrier to entry, weaken interoperability, and leave much potentially reusable knowledge confined to publications or single-case applications, preventing its conversion into cross-model capabilities. Process equations derived from experiments or data analysis are often left unused because researchers lack the computational expertise to implement and integrate them into modeling systems. Even when implementations are mature, reproducing results across models requires substantial effort to learn their unique structures and workflows. This, in turn, hinders cross-regional and cross-scale model validation and application.

To address these limitations, Phase II moves beyond reliance on a single process-based model toward a more generalizable and adaptable modeling paradigm. One possible approach is to encode watershed-specific attributes as embedding vectors and use them as auxiliary inputs during training to improve robustness across regions and scenarios. However, this strategy depends on case-specific scenario simulations for training data, which involve high technical and computational costs, including data acquisition, inputs preprocessing, parameter configuration, and calibration *(Clark et al., 2015, 2017; Hutton et al., 2016)*. More importantly, it does not fundamentally overcome the structural constraints of existing process models.

Our core strategy in Phase II is to extract knowledge directly from process equations. By abstracting and standardizing the core mechanisms and structural equations of different process-based models, we can develop a unified, model-agnostic framework. Leveraging advances in GPU-accelerated computing and deep learning architectures, these process-based insights can be modularized into reusable neural network components, assembled into an extensible knowledge base, and integrated into flexible deep learning frameworks to enable cross-model, cross-regional knowledge transfer and structural reconfiguration.

### 3.2    A Flexible and Generalizable Graph-Based Framework for Ecohydrological Modeling

To implement the strategy outlined above, it is essential to standardize the core structural units and common process representations that underlie ecohydrological models. Most models, whether lumped, semi-distributed, or fully distributed, can be viewed as coupled systems of internally consistent process equations. Their core lies in the parameterized simulation of ecohydrological processes within spatial units, while the process equations define modular structures that govern dynamic updates and maintain physical consistency.

Building on this concept, we propose a GNN-based framework (**Fig. 4**) that abstracts spatial heterogeneity and multi-process coupling through graph structures, enabling structural generalization and interoperability across spatial resolutions and structural complexities. In this framework, nodes represent spatial units with multidimensional attributes such as climate, soil, land use, and topography, while edges describe interactions including flow pathways, material transport, and ecological coupling. Specifically:

- A graph without edges corresponds to lumped models (e.g., HBV), where process equations describe the integrated system response.
- A multi-node directed graph corresponds to semi-distributed models (e.g., SWAT), where nodes represent HRUs or sub-basins, and edges represent ecohydrological connectivity between upstream and downstream units.
- A high-resolution grid graph corresponds to fully distributed models (e.g., VELMA), where topology is derived from flow direction networks (e.g., DEM-based) to explicitly capture fine-scale processes and spatial heterogeneity.

Graph structures can be transformed through graph coarsening (e.g., via node selection or clustering) and uncoarsening (e.g., via node replication or interpolation) to support multi-scale modeling. At each time step, information from neighboring nodes is "aggregated" (using various relevant operators such as sum, mean, max, or other custom-created ones) and used to update the target node state via: (1) embedded process equations, (2) deep learning layers, or (3) hybrid process–ML methods. The general form is:

$$h_v^{(t+1)} = UPDATE\left(h_v^{(t)}, AGGREGATE\left(\left\{h_u^{(t)} \mid u \in \mathcal{N}(v)\right\}\right)\right)$$

Here, $h_v^{(t)}$ is the feature representation of node $v$ at iteration $t$, including both its internal state and external forcing variables. $\mathcal{N}(v)$ is the set of neighbors. The $AGGREGATE$ integrates neighbor information, and $UPDATE$ combines it with the current state to produce the new state.

This aggregate–update cycle enables efficient information propagation, capturing multi-scale dependencies. The framework supports embedding process equations into ML architectures and gradient-based parameter optimization, overcoming the rigidity of traditional models and improving calibration efficiency. Its graph-based design also facilitates integration of heterogeneous data sources such as remote sensing, forecasts, and sensor networks via node-level embeddings, combining physical consistency with data-driven adaptability. With modular composition and end-to-end trainability, it provides a unified and scalable foundation for ecohydrological modeling and data-driven decision support.

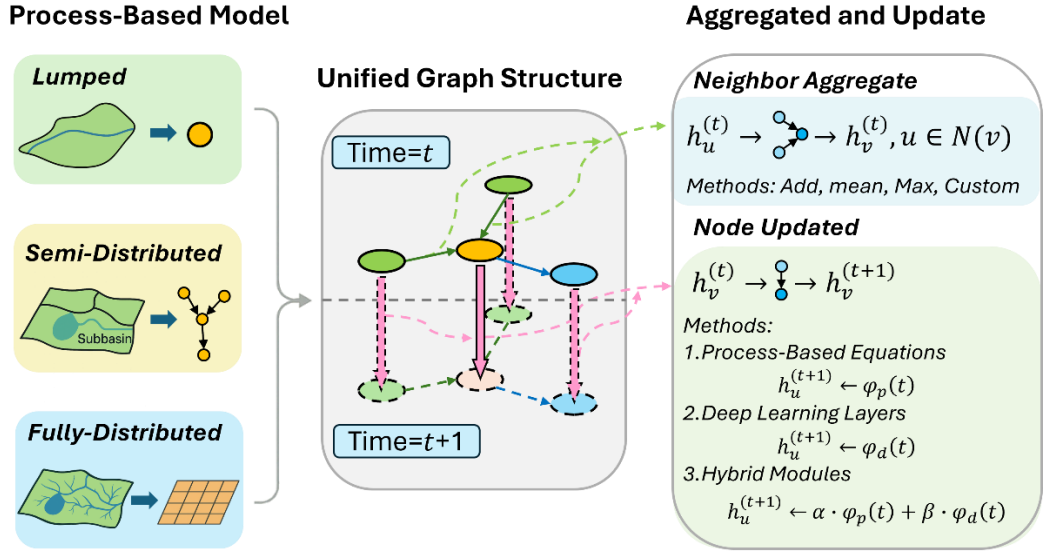

*Fig. 4 A unified graph-based framework for ecohydrological modeling integrating process-based mechanisms, machine learning, and spatial heterogeneity.*

### 3.3 Implementation of EcoHydroModel

Based on the framework, we developed **EcoHydroModel** (**Fig. 5**), a GNN-based toolkit for ecohydrological modeling implemented in PyTorch. The system is unified, modular, and extensible. It supports seamless coupling of physical process equations with ML models in an end-to-end trainable system. The central controller coordinates four core modules: *DataManager, Updater, Trainer, and Visualizer*. Source code and documentation are publicly available on GitHub (https://github.com/jlonghku/EcoHydroModel).

The framework is highly modular, allowing users to extend the *Updater* class to flexibly define ecohydrological processes. Verified model instances, established models, and process-equation knowledge from literature can be converted into reusable components within a module pool by following the provided model specifications, either manually or with assistance from large language models (LLMs) such as ChatGPT or Claude. The *Updater* supports three strategies (i.e., process-based, ML, and hybrid) which can be freely combined and extended as needed. For example, water flow can be modeled with a linear reservoir, the kinematic wave equation, or the full Saint-Venant equations. Processes with limited data, such as evapotranspiration or nutrient cycling, can be modeled with neural networks. In hybrid mode, physical equations can govern soil moisture dynamics, while ML modules correct energy fluxes in heterogeneous vegetation zones. This flexibility preserves physical interpretability while adapting to diverse contexts.

To balance GPU parallelism with spatiotemporal physical dependencies, the framework provides three configurable update strategies: (1) parallel synchronous updating, suited for short time steps or weakly coupled processes (e.g., vegetation biomass); (2) topology-based asynchronous updating, suited for longer time steps or strongly coupled processes (e.g., surface–groundwater interactions); and (3) convergence-based iterative updating, suited for scenarios with high accuracy requirements.

The *Trainer* module enables fully differentiable training and efficient parameter calibration, grounded in a generalized knowledge distillation framework that links diverse sources and targets. It unifies ML-based distillation (Data, Process → ML) with traditional calibration and cross-model transfer, supporting four typical modes:

1. **Data-driven Calibration (Data-to-Process Distillation)**: Observational data are used to train process-based models to approximate real system behavior.

2. **Data-driven Model Training (Data-to-ML Distillation)**: Observational data are used to directly guide end-to-end training of ML models to learn system mappings.
3. **Process-Guided Model Extraction (Process-to-ML Distillation)**: Behavioral patterns encoded by high-fidelity process models are distilled into ML surrogates.
4. **Cross-Model Knowledge Transfer (Process-to-Process Distillation)**: Behavioral knowledge from one process model/equation is transferred to another with a different structure, enabling cross-structural reuse.

For spatial consistency, the *DataManager* implements graph coarsening to switch between lumped, semi-distributed, and fully distributed structures, enabling unified multi-scale modeling. The Visualizer generates time-series plots, performance metrics, and spatial visualizations. By combining process-based and data-driven components at the node level, **EcoHydroModel** provides a unified platform for integrating and comparing existing ecohydrological models, maintaining structural consistency and scalability across spatial resolutions and modeling paradigms.

### 3.4 Simple Case Demonstration

**Figs. 5(a)–(e)** present a fully distributed case study for the Samish watershed, comparing three modeling strategies under identical inputs. The GNN-based EcoHydroModel framework integrates nitrification process equations with multilayer perceptron (MLP) modules, enabling performance comparison across strategies. Details of process equations and model configurations are provided in the supplementary materials.

**Figs. 5(a)** and **5(b)** show the spatial distribution of average nitrification from the original VELMA and EcoHydroModel, both using the same process-based nitrification equations *(Del Grosso et al., 2000, 2006)*. Results show high spatial consistency, confirming that EcoHydroModel can accurately reproduce traditional model outputs when supplied with equivalent equations and parameters. This validates both its physical consistency and its capacity to inherit existing formulations.

**Figs. 5(c)–(e)** compare mean NSE–KGE scores over the entire simulation period for three strategies:

1. **Process-based model** *(Parton et al., 2001)* (**Fig. 5(c)**): Despite using a different nitrification equation, the model reproduced outputs with good accuracy, demonstrating transferability between formulations in EcoHydroModel. Its modular design allows flexible replacement or combination of equations, enabling direct comparison and integration across mechanistic assumptions.
2. **Pure ML model** (MLP, 4 layers × 64 units) (**Fig. 5(d)**): Overall accuracy was slightly higher than the simplified process-based model, with particularly strong performance in high-value regions, reflecting its ability to capture complex spatiotemporal patterns. However, stability in low-value regions declined, and the absence of process constraints risks error accumulation in long simulations.
3. **Hybrid model** (**Fig. 5(e)**): By combining the trend-capturing capability of the process module with the residual refinement of the ML module, the hybrid approach achieved the best spatial predictions of nitrification, maintaining both physical consistency and adaptive precision.

Overall, EcoHydroModel offers structural flexibility and cross-model interoperability. Especially when process equations capture the dominant trends, the framework can integrate deep learning components to characterize implicit nonlinear relationships that process equations fail to capture. This capability makes EcoHydroModel a powerful tool for scalable, high-accuracy, and computationally efficient simulations in complex, heterogeneous watersheds.

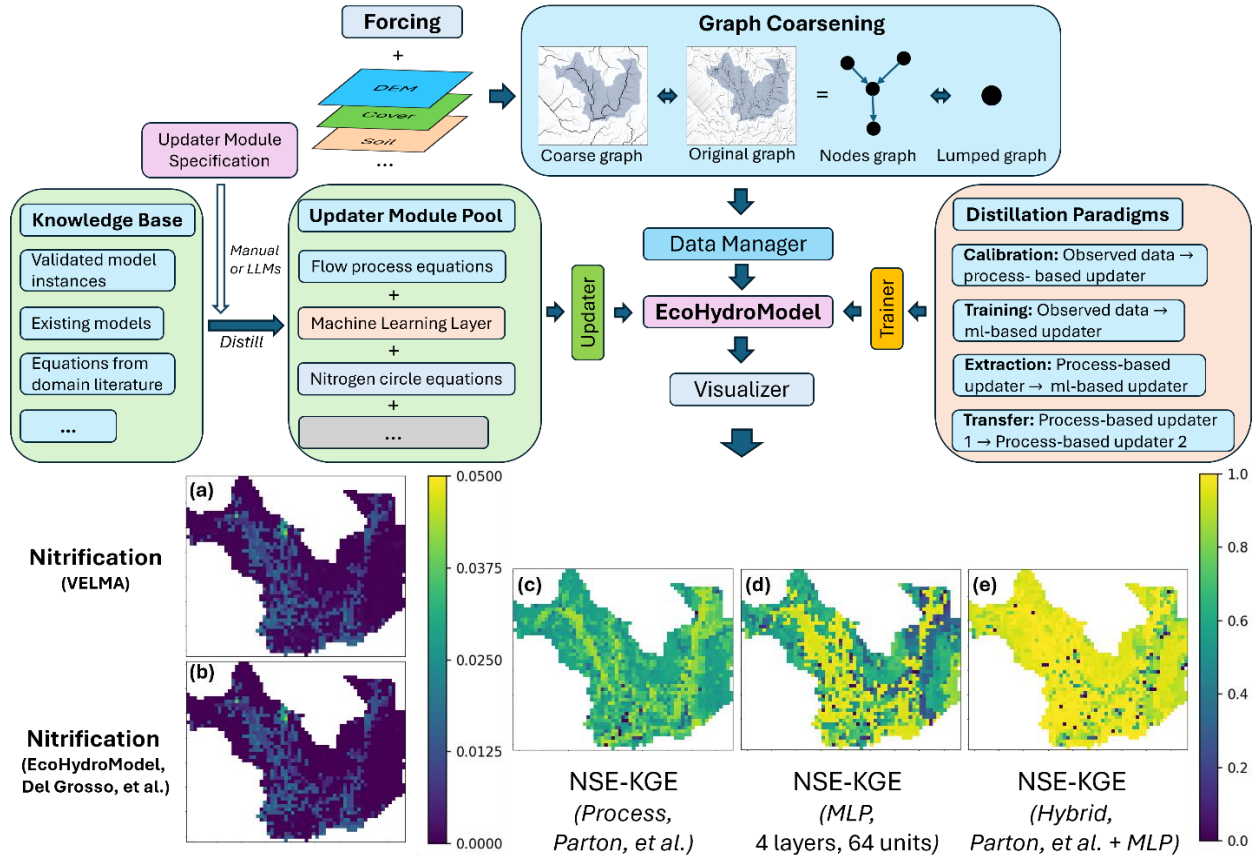

*Fig. 5 Implementation of EcoHydroModel, a unified GNN-based toolkit for ecohydrological modeling. (a), (b) show the mean nitrification ($g\,m^{-2}\,day^{-1}$) simulated by (a) the original VELMA model and (b) the unified framework using the equation from Del Grosso et al. (c)–(e) compare spatial performance of different modeling strategies, quantified as the mean of NSE-KGE scores: (c) simplified process-based model (Parton equation), (d) pure ML model (MLP), and (e) hybrid model integrating process-based and ML components. Warmer (yellower) colors indicate higher spatial agreement.*

## 4 Phase III: Cognitive Distillation

### 4.1 Toward Structural Distillation: AI-Driven Expert Modeling

Although Phase II establishes a unified structural framework and begins to integrate ML beyond simply learning/correcting the process equation representations, the deep fusion of process models and AI remains incomplete. Many process equations originate in laboratory experiments or small-scale field studies and are subsequently refined through expert reasoning and accumulated domain experience. As a result, domain knowledge is inherently dynamic, evolving through the interplay between empirical evidence and scientific cognition.

However, the capacity of individual modelers is inherently limited, and updates often strain our available expertise. Incorporating new observations or previously unrepresented processes may improve data agreement but can also induce unrealistic system behavior, requiring repeated structural reviews to identify hidden flaws. This highlights uncertainties in understanding deeper mechanisms *(Clark et al., 2011)*. Even under identical forcings, models can differ markedly in the magnitude and functional form of key processes, such as groundwater recharge *(Gnann et al., 2023)*. This partly reflects the need for exceptional integrative expertise to synthesize knowledge across cases. These constraints can also lead to incomplete or inaccurate reasoning, especially for non-experts *(Gupta et al., 2012)*. Moreover, human decision-making is influenced by path dependence (i.e., the tendency to follow established choices and practices), with model selection often shaped more by precedent than by scientific suitability *(Addor & Melsen, 2019)*.

Advances in AI offer a pathway to address these challenges. Rather than relying solely on static domain knowledge, AI could be tasked with: (i) learning the cognitive frameworks that experts use in model construction, calibration, and refinement, (ii) emulating their reasoning, and (iii) adapting through continuous interaction with multi-source data. The central goal of this phase is to deconstruct domain knowledge and expert logic into learnable, parameterizable components, and then embed them into AI-driven modeling frameworks to create an evolving, expert-level modeling intelligence. Although still at a conceptual stage of development, we contend that such a cognitive-framework-centered transformation could redefine the paradigm of next-generation process models. The following section discusses the limitations of current AI systems and the technical and conceptual challenges that must be overcome to realize this vision.

### 4.2 Why AI Still Struggles to Emulate Expert Use of Ecohydrological Models

Despite rapid advances in AI, particularly the exceptional performance of LLMs in general tasks such as question answering, text generation, and logical reasoning, their application in ecohydrological modeling remains limited. Ecohydrological systems involve complex physical mechanisms, heterogeneous multi-source data, explicit spatial structures, and tightly coupled process feedbacks, which lie beyond the scope of most current AI learning paradigms. The key challenges fall into four levels: data, model, reasoning, and application.

1. **Data level**
    - *Lack of domain semantic understanding*: Current AI models struggle to fully comprehend the contextual meanings of variables, physical units, governing equations, and feedback mechanisms in ecohydrological systems. This limits their ability to configure models, identify dominant processes, or implement structural changes.
    - *Limited accessibility of data and code*: Many ecohydrological models developed in academic or government contexts lack open interfaces, comprehensive documentation, and complete metadata. Commercial models are better documented but usually restrict source code access. The absence of standardized, well-structured code constrains AI's ability to perform semantic parsing, modular execution, and behavioral learning.

2. **Model level**
    - *High structural complexity*: Models such as VELMA, RHESSys, and MIKE SHE integrate hydrological, biogeochemical, and ecological processes within high-dimensional parameter and state spaces, heterogeneous input formats, and intricate module dependencies. This complexity manifests as unstructured data and code architectures, hindering AI from constructing coherent, structured representations.

3. **Reasoning level**
    - *Limited domain reasoning and interpretability*: Current AI systems remain limited in their ability to perform causal, spatiotemporal, and mechanistic reasoning. For example, diagnosing the drivers of increased nitrate loss or tracing the origins of model biases requires constructing a coherent reasoning chain from observations, through process understanding, to model behavior. This is an ability that AI still lacks in ecohydrology.
    - *Implicit expert workflows*: Traditional modeling involves tasks such as data preprocessing, parameter configuration, model execution, and post-processing analysis. These steps rely heavily on expert judgment and are seldom formalized in a machine-interpretable form, making it challenging for AI systems to replicate or learn the underlying logic.

4. **Application level**
    - *Limited generalizability and transferability*: Ecohydrological modeling depends strongly on regional characteristics, climatic conditions, land-use patterns, and model configurations. Without domain-specific priors, AI models often struggle to transfer knowledge across watersheds or adapt to unfamiliar simulation settings.
    - *Lack of standardized benchmarks and evaluation systems*: The absence of widely accepted benchmark datasets, reference models, and unified protocols hampers systematic evaluation of AI approaches. Without such standards, it is difficult to fairly compare their generality, robustness, and practical value under reproducible and scalable conditions.

## 4.3 EBHM: A Cognitive Analogy for Future AI-Driven Expert Modeling

To address the limitations of traditional ecohydrological modeling in knowledge integration, cross-task generalization, and interpretability, we propose the EBHM (Eyes–Brain–Hands–Mouth), an expert-emulating cognitive architecture (**Fig. 6**). It emulates the full modeling workflow of human experts, from perception and acquisition of information to communication of results, and establishes a continuous feedback learning loop (E→B→H→M→E) that lays the foundation for AI-driven, expert-level ecohydrological modeling.

**Eyes (E): Perception and Knowledge Acquisition**

1. **Multi-source data fusion and scenario adaptation**: Observational datasets, model repositories, and domain literature are integrated via retrieval-augmented generation (RAG) and few-shot learning, enhancing AI's semantic understanding of model structures, watershed characteristics, and modeling objectives.
2. **Model source code parsing and interface abstraction**: Code-focused LLMs parse I/O schemas, parameter definitions, and module dependencies from process-based model code, converting them into callable APIs and specifications to lower barriers for AI-assisted configuration and deployment.

**Brain (B): Knowledge Integration and Human-like Reasoning**

1. **Structured knowledge graphs and instruction tuning**: Structured knowledge graphs encode variable definitions, process linkages, and parameter pathways. Instruction tuning aligns this knowledge with domain semantics for targeted reasoning.
2. **Heuristic reasoning**: Tree-of-thoughts and ReAct (Reasoning + Acting) approaches decompose complex problems into actionable steps. Hypothesis testing and feedback-driven optimization refine decision paths, yielding transparent, traceable reasoning chains/trees for expert-like diagnosis, structural optimization, and parameter calibration.
3. **Continuous learning and adaptation**: Memory-augmented networks and transformer-based retrieval modules store and recall modeling experience, enabling cross-task transfer and long-term adaptation.

**Hands (H): Model Construction, Execution, and Validation**

1. **Multi-agent workflow system**: AI agents collaborate across the full pipeline of model construction and execution. They begin with the definition of conceptual structures and the assembly or selection of model components. The workflow then proceeds through data preprocessing and integration, parameter configuration, and simulation monitoring, and concludes with result analysis and validation. The Model Context Protocol (MCP) ensures semantic coordination and modular management, improving automation and flexibility.
2. **Goal-oriented modeling and generative planning**: AI agent autonomously sets objectives and develops scenario plans (e.g., land use change, climate stress, agricultural interventions).

Integration of knowledge graphs and causal reasoning enables evaluation of ecohydrological impacts and supports multi-objective optimization.

**Mouth (M): Result Interpretation and Communication**

1. **Multimodal output and interpretability**: Natural language summaries, visual analytics, and automated plotting provide process explanations, sensitivity analyses, and time-series visualizations, improving result interpretability and communication.
2. **Reasoning playback and oversight**: Reasoning chains and configurations are logged for transparency, reproducibility, and auditability. Rule-based oversight, post-hoc evaluation, and human review strengthen reliability and trust.

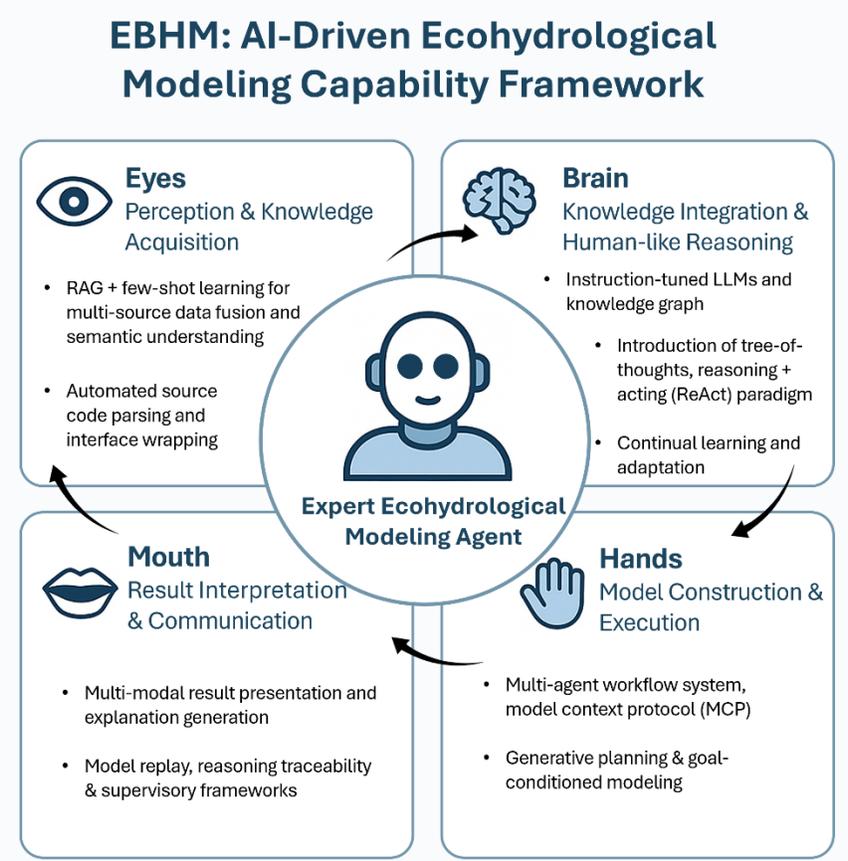

*Fig. 6 The EBHM framework: An expert-emulating AI architecture for ecohydrological modeling*

### 4.4　Illustrative Scenario: Future Expert-Agent Workflow

To demonstrate the application potential of the proposed architecture, we present a forward-looking expert–agent workflow for nitrogen load mitigation in the Samish watershed. In this scenario, an AI-driven expert ecohydrological modeling agent, built on the EBHM, collaborates with researchers to complete the full modeling cycle from task definition to solution optimization.

**Micro-case: Future Expert–Agent Workflow for Nitrogen Load Mitigation**

- **Objective** – Assess and optimize measures to reduce nitrogen export from the mainstem to downstream receiving waters.

- **Task initiation** – The agent engages in interactive dialogue to clarify objectives, identify evaluation metrics, and define output formats. It drafts a workflow grounded in domain knowledge and confirms an execution blueprint with the user.
- **Perception phase (E)** – The agent retrieves and integrates multi-source data, including NOAA meteorological and USGS streamflow records, state nitrogen monitoring data, and land use and soil type maps. It performs standardized preprocessing and quality control to ensure consistency and reliability.
- **Reasoning phase (B)** – Using knowledge graphs and prior cases, the agent generates an initial portfolio of mitigation strategies, such as 15–30 m riparian buffers, wetland restoration covering 2–5% of the watershed, and optimized fertilizer application timing for cover crops. It predicts process responses, tests hypotheses, and adjusts reasoning pathways as needed.
- **Execution phase (H)** – The agent first establishes a conceptual model informed by prior domain knowledge and cognitive frameworks, and then adaptively invokes process models or assembles relevant EcoHydroModel modules (e.g., runoff generation, nitrogen cycling, river transport) according to the mitigation strategies. It runs batch simulations at appropriate resolutions and evaluates multiple objectives, including nitrogen reduction magnitude and cost-effectiveness. If results meet targets, the process moves to communication; if not, the agent refines strategies based on model responses and re-runs simulations, forming a closed-loop optimization cycle.
- **Communication phase (M)** – The agent presents results via scenario tables, time-series plots, and spatial maps, accompanied by the reasoning chain/tree and model configuration for transparency and reproducibility. It then discusses next steps with the user for further optimization.

This illustrative case shows how the EBHM framework can guide an AI-driven expert agent through the complete modeling loop—from perception to reasoning, execution, and communication—within a real ecohydrological context. With continued enhancement of EBHM capabilities, AI can co-evolve with process-based ecohydrological models to form an integrated intelligent modeling system.

## 5 Discussion

### 5.1 Recognizing Overfitting and Overparameterization

In the **behavioral distillation** phase, we employed ML models to capture the complex input–output dynamics of ecohydrological process-based models. Although deep neural networks may internalize region-specific features, causing apparent overfitting and reduced cross-region generalization, *the primary limitation does not stem from the deep learning models themselves*. Evidence shows that the strong representational capacity of deep learning networks, coupled with the benign overfitting, allows them to maintain high generalization performance even when overfitting occurs *(Bartlett et al., 2020)*. In most cases, *the real bottleneck is training data generation*: computationally intensive and structurally diverse process-based models often produce training data from a single region or scenario, limiting the applicability to new contexts. Under these constraints, the primary role of Phase I is to enhance traditional models within a given region or scenario. It is efficient and simple, sometimes trainable on a single time series, without requiring cross-scenarios or cross-spatial generalization.

In the **structural distillation** phase, we aim to distill process equations from established model instances, mature process models, and domain equations, progressively embedding them into ML models. A common concern from traditional model users is: if process equations are already integrated into the unified EcoHydroModel, why transfer them into ML models with large parameter capacity? Does this not risk "overparameterization"?

In traditional process-based models, excessive parameters increase calibration burdens and require simultaneous agreement with both physical plausibility and observations, making optimization complex and expertise-dependent. If structural assumptions are biased, calibration results may remain unsatisfactory. Deep learning differs fundamentally: *large parameter counts do not necessarily cause overfitting and may even improve the chance of finding simple, generalizable solutions*. In high-dimensional parameter spaces, broad flat minima with low curvature occupy larger volumes, making them easier to locate and inherently more robust *(Huang et al., 2020; Maddox et al., 2020)*. Large models can explore wider hypothesis spaces, and optimization naturally tends to favor structurally simple and compressible solutions, avoiding the risks associated with sharp, complex minima *(Alexander et al., 2025; Geiping et al., 2022)*.

Moreover, greater parameterization also improves mode connectivity, allowing multiple local optima to be linked by low-loss paths within the same flat basin *(Wilson, 2025)*. This supports the discovery of robust solutions and enhances stability to parameter perturbations. In contrast, traditional process models, with fewer parameters and rigid constraints, often have optima separated by high-energy barriers, requiring computationally expensive global optimization.

Our discussion of overfitting and overparameterization is not intended to prescribe specific technical solutions but rather to ease the concerns of model users who may not have a comprehensive ML background. Large parameter counts and overfitting are not inherently problematic. In most cases, large models naturally converge toward solutions that both fit the data and are compressible. When needed, such solutions can be further refined through lightweight soft inductive biases, such as moderate regularization, stochastic weight averaging, or gradient penalties. In the future, such capabilities are likely to be embedded directly into ML toolkits, allowing users of traditional ecohydrological models to benefit without mastering the underlying technical details.

## 5.2 Cross-Domain Adaptability and Extension Potential

The proposed three-phase framework is demonstrated through ecohydrological process modeling, but it is broadly applicable to other process-based domains. Specifically, it adapts to systems with clearly defined driving variables, analyzable state evolution mechanisms, and explicit or implicit spatial or logical topologies.

In the **behavioral distillation** phase, simplified model variants are created through resolution coarsening or process reduction, and their input–output dynamics are enhanced using ML. This principle readily extends to domains such as atmospheric circulation, land–atmosphere interactions, and regional ocean dynamics, where traditional models are often high-resolution, computationally expensive, and structurally complex. Distilled models derived from such simplifications can support rapid calibration, scenario analysis, and decision-making.

In the **structural distillation** phase, process equations and data flows are abstracted as graph representations. This approach applies equally well to domains like urban water systems, river networks, transportation flows, and power grids, which exhibit well-defined spatial or logical topology. By identifying governing processes such as flow, heat, density, or pressure and abstracting them into graph form, existing process equations can be embedded into AI architectures as reusable components, and integrated into a unified, interoperable modeling framework.

In the **cognitive distillation** phase, expert knowledge is embedded within AI systems. This knowledge extends beyond equations to include heuristics, modeling strategies, diagnostic procedures, and decision rules shared across domains. EBHM architecture formalizes such expertise into AI components, enabling expert-level workflows through a closed-loop cycle of perception, reasoning, execution, and communication. Its capabilities are broadly transferable, and by integrating open-source foundation models such as Llama 3 and Mistral and applying domain-specific fine-tuning, EBHM can evolve into expert-level agents across diverse process-based domains.

In summary, the concepts and methods of behavioral, structural, and cognitive distillation offer strong cross-domain adaptability. With appropriate customization to sector-specific requirements, they enable efficient simulation, structural unification, and expert-level intelligence in process modeling, providing a robust theoretical and technical foundation for intelligent modeling across diverse disciplines.

# 6 Conclusion

This study presents a unified three-phase framework for integrating AI with process-based ecohydrological models, grounded in progressive knowledge distillation: (1) behavioral distillation for targeted model enhancement, (2) structural distillation through graph-based abstraction of process mechanisms, and (3) cognitive distillation to embed expert reasoning into AI agents.

Using the Samish watershed as a case study we have shown how these phases link physical interpretability with adaptive intelligence, providing a pathway from efficiency-oriented augmentation to autonomous, expert-level modeling systems.

In **Phase I** (behavioral distillation) computational efficiency and predictive accuracy were improved within specific scenarios. ML surrogates were trained on high-fidelity process model outputs to replicate daily streamflow and nitrate ($NO_3$) loss dynamics. This enabled rapid gradient-based calibration and scenario evaluation without requiring broad spatial or temporal generalization. Hybrid strategies combining residual and transfer learning achieved an effective balance between accuracy and computational cost.

In **Phase II** (structural distillation), process mechanisms were reformulated into modular, graph-based deep learning architectures, enabling interoperability across models, multi-scale representation, and physically consistent hybrid learning. Domain equations, such as nutrient cycling, were abstracted into a unified GNN-based framework that preserves spatial heterogeneity and mechanistic relationships. By embedding process modules into a flexible, scalable architecture, the framework integrates diverse environmental components while maintaining their physical grounding.

Finally, we discussed how via **Phase III** (cognitive distillation), the framework could be extended toward autonomous modeling by embedding expert cognitive structures into AI agents. Using the EBHM (Eyes–Brain–Hands–Mouth) architecture, we illustrated a future expert-agent workflow for nitrogen reduction planning, showing how cognitive distillation can transform AI from a passive computational tool into an active, collaborative modeling partner. Developmental work towards this outcome is currently in progress.

In summary, the framework offers both a practical roadmap for enhancing ecohydrological models in targeted contexts and a conceptual foundation for developing intelligent, physically grounded, expert-emulating systems, with potential applicability to other process-based modeling domains.

# 7 Acknowledgements

This work was supported by the Allen Family Philanthropies.

# 8 References

Addor, N., & Melsen, L. A. (2019). Legacy, Rather Than Adequacy, Drives the Selection of Hydrological Models. *Water Resources Research*, *55*(1), 378–390. https://doi.org/10.1029/2018WR022958

Aieb, A., Liotta, A., Jacob, A., Ferrario, I. F., & Yaqub, M. A. (2025). An Innovative Approach for Calibrating Hydrological Surrogate Deep Learning Models. *Remote Sensing*, *17*(11), 1916. https://doi.org/10.3390/rs17111916

Alexander, Y., Slutzky, Y., Ran-Milo, Y., & Cohen, N. (2025). *Do Neural Networks Need Gradient Descent to Generalize? A Theoretical Study* (No. arXiv:2506.03931). arXiv. https://doi.org/10.48550/arXiv.2506.03931

Banda, V. D., Dzwairo, R. B., Singh, S. K., & Kanyerere, T. (2022). Hydrological Modelling and Climate Adaptation under Changing Climate: A Review with a Focus in Sub-Saharan Africa. *Water*, *14*(24), 4031. https://doi.org/10.3390/w14244031

Bartlett, P. L., Long, P. M., Lugosi, G., & Tsigler, A. (2020). Benign overfitting in linear regression. *Proceedings of the National Academy of Sciences*, *117*(48), 30063–30070. https://doi.org/10.1073/pnas.1907378117

Beven, K. J. (2018). On hypothesis testing in hydrology: Why falsification of models is still a really good idea. *WIREs Water*, *5*(3), e1278. https://doi.org/10.1002/wat2.1278

Bonetti, S., Wei, Z., & Or, D. (2021). A framework for quantifying hydrologic effects of soil structure across scales. *Communications Earth & Environment*, *2*(1), 107. https://doi.org/10.1038/s43247-021-00180-0

Brewer, S. K., Worthington, T. A., Mollenhauer, R., Stewart, D. R., McManamay, R. A., Guertault, L., & Moore, D. (2018). Synthesizing models useful for ecohydrology and ecohydraulic approaches: An emphasis on integrating models to address complex research questions. *Ecohydrology*, *11*(7), e1966. https://doi.org/10.1002/eco.1966

Clark, M. P., Bierkens, M. F. P., Samaniego, L., Woods, R. A., Uijlenhoet, R., Bennett, K. E., Pauwels, V. R. N., Cai, X., Wood, A. W., & Peters-Lidard, C. D. (2017). The evolution of process-based hydrologic models: Historical challenges and the collective quest for physical realism. *Hydrology and Earth System Sciences*, *21*(7), 3427–3440. https://doi.org/10.5194/hess-21-3427-2017

Clark, M. P., Kavetski, D., & Fenicia, F. (2011). Pursuing the method of multiple working hypotheses for hydrological modeling. *Water Resources Research*, *47*(9). https://doi.org/10.1029/2010WR009827

Clark, M. P., Nijssen, B., Lundquist, J. D., Kavetski, D., Rupp, D. E., Woods, R. A., Freer, J. E., Gutmann, E. D., Wood, A. W., Brekke, L. D., Arnold, J. R., Gochis, D. J., & Rasmussen, R. M. (2015). A unified approach for process-based hydrologic modeling: 1. Modeling concept. *Water Resources Research*, *51*(4), 2498–2514. https://doi.org/10.1002/2015WR017198

Clement, J. J. (2022). Multiple Levels of Heuristic Reasoning Processes in Scientific Model Construction. *Frontiers in Psychology*, *13*. https://doi.org/10.3389/fpsyg.2022.750713

Dedhia, B., Kansal, Y., & Jha, N. K. (2025). *Bottom-up Domain-specific Superintelligence: A Reliable Knowledge Graph is What We Need* (No. arXiv:2507.13966). arXiv. https://doi.org/10.48550/arXiv.2507.13966

Del Grosso, S. J., Parton, W. J., Mosier, A. R., Ojima, D. S., Kulmala, A. E., & Phongpan, S. (2000). General model for $N_2O$ and $N_2$ gas emissions from soils due to dentrification. *Global Biogeochemical Cycles*, *14*(4), 1045–1060. https://doi.org/10.1029/1999GB001225

Del Grosso, S. J., Parton, W. J., Mosier, A. R., Walsh, M. K., Ojima, D. S., & Thornton, P. E. (2006). DAYCENT National-Scale Simulations of Nitrous Oxide Emissions from Cropped Soils in the United States. *Journal of Environmental Quality*, *35*(4), 1451–1460. https://doi.org/10.2134/jeq2005.0160

Eythorsson, D., & Clark, M. (2025). Toward Automated Scientific Discovery in Hydrology: The Opportunities and Dangers of AI Augmented Research Frameworks. *Hydrological Processes*, *39*(1), e70065. https://doi.org/10.1002/hyp.70065

Fatichi, S., Vivoni, E. R., Ogden, F. L., Ivanov, V. Y., Mirus, B., Gochis, D., Downer, C. W., Camporese, M., Davison, J. H., Ebel, B., Jones, N., Kim, J., Mascaro, G., Niswonger, R., Restrepo, P., Rigon, R., Shen, C., Sulis, M., & Tarboton, D. (2016). An overview of current applications, challenges, and future trends in distributed process-based models in hydrology. *Journal of Hydrology*, *537*, 45–60. https://doi.org/10.1016/j.jhydrol.2016.03.026

Feldman, A. F., Short Gianotti, D. J., Dong, J., Akbar, R., Crow, W. T., McColl, K. A., Konings, A. G., Nippert, J. B., Tumber-Dávila, S. J., Holbrook, N. M., Rockwell, F. E., Scott, R. L., Reichle, R. H., Chatterjee, A., Joiner, J., Poulter, B., & Entekhabi, D. (2023). Remotely Sensed Soil Moisture Can Capture Dynamics Relevant to Plant Water Uptake. *Water Resources Research*, *59*(2), e2022WR033814. https://doi.org/10.1029/2022WR033814

Garzón, A., Kapelan, Z., Langeveld, J., & Taormina, R. (2022). Machine Learning-Based Surrogate Modeling for Urban Water Networks: Review and Future Research Directions. *Water Resources Research*, *58*(5), e2021WR031808. https://doi.org/10.1029/2021WR031808

Geiping, J., Goldblum, M., Pope, P. E., Moeller, M., & Goldstein, T. (2022). *Stochastic Training is Not Necessary for Generalization* (No. arXiv:2109.14119). arXiv. https://doi.org/10.48550/arXiv.2109.14119

Gnann, S., Reinecke, R., Stein, L., Wada, Y., Thiery, W., Müller Schmied, H., Satoh, Y., Pokhrel, Y., Ostberg, S., Koutroulis, A., Hanasaki, N., Grillakis, M., Gosling, S. N., Burek, P., Bierkens, M. F. P., & Wagener, T. (2023). Functional relationships reveal differences in the water cycle representation of global water models. *Nature Water*, *1*(12), 1079–1090. https://doi.org/10.1038/s44221-023-00160-y

Gupta, H. V., Clark, M. P., Vrugt, J. A., Abramowitz, G., & Ye, M. (2012). Towards a comprehensive assessment of model structural adequacy. *Water Resources Research*, *48*(8). https://doi.org/10.1029/2011WR011044

Huang, W. R., Emam, Z., Goldblum, M., Fowl, L., Terry, J. K., Huang, F., & Goldstein, T. (2020). *Understanding Generalization Through Visualizations*. 87–97. https://proceedings.mlr.press/v137/huang20a.html

Hutton, C., Wagener, T., Freer, J., Han, D., Duffy, C., & Arheimer, B. (2016). Most computational hydrology is not reproducible, so is it really science? *Water Resources Research*, *52*(10), 7548–7555. https://doi.org/10.1002/2016WR019285

Karpatne, A., Atluri, G., Faghmous, J. H., Steinbach, M., Banerjee, A., Ganguly, A., Shekhar, S., Samatova, N., & Kumar, V. (2017). Theory-Guided Data Science: A New Paradigm for Scientific Discovery from Data. *IEEE Transactions on Knowledge and Data Engineering*, *29*(10), 2318–2331. https://doi.org/10.1109/TKDE.2017.2720168

Klotz, D., Kratzert, F., Gauch, M., Keefe Sampson, A., Brandstetter, J., Klambauer, G., Hochreiter, S., & Nearing, G. (2022). Uncertainty estimation with deep learning for rainfall–runoff modeling. *Hydrology and Earth System Sciences*, *26*(6), 1673–1693. https://doi.org/10.5194/hess-26-1673-2022

Knoben, W. J. M., & Spieler, D. (2022). Teaching hydrological modelling: Illustrating model structure uncertainty with a ready-to-use computational exercise. *Hydrology and Earth System Sciences*, *26*(12), 3299–3314. https://doi.org/10.5194/hess-26-3299-2022

Kratzert, F., Klotz, D., Shalev, G., Klambauer, G., Hochreiter, S., & Nearing, G. (2019). Towards learning universal, regional, and local hydrological behaviors via machine learning applied to large-sample datasets. *Hydrology and Earth System Sciences*, *23*(12), 5089–5110. https://doi.org/10.5194/hess-23-5089-2019

Kratzert, F., Nearing, G., Addor, N., Erickson, T., Gauch, M., Gilon, O., Gudmundsson, L., Hassidim, A., Klotz, D., Nevo, S., Shalev, G., & Matias, Y. (2023). Caravan—A global community dataset for large-sample hydrology. *Scientific Data*, *10*(1), 61. https://doi.org/10.1038/s41597-023-01975-w

Maddox, W. J., Benton, G., & Wilson, A. G. (2020). *Rethinking Parameter Counting in Deep Models: Effective Dimensionality Revisited* (No. arXiv:2003.02139). arXiv. https://doi.org/10.48550/arXiv.2003.02139

Maity, R., Srivastava, A., Sarkar, S., & Khan, M. I. (2024). Revolutionizing the future of hydrological science: Impact of machine learning and deep learning amidst emerging explainable AI and transfer learning. *Applied Computing and Geosciences*, *24*, 100206. https://doi.org/10.1016/j.acags.2024.100206

Massoud, E. C., Hoffman, F., Shi, Z., Tang, J., Alhajjar, E., Barnes, M., Braghiere, R. K., Cardon, Z., Collier, N., Crompton, O., Dennedy-Frank, P. J., Gautam, S., Gonzalez-Meler, M. A., Green, J. K., Koven, C., Levine, P., MacBean, N., Mao, J., Mills, R. T., … Zarakas, C. (2023). *Perspectives on Artificial Intelligence for Predictions in Ecohydrology*. https://doi.org/10.1175/AIES-D-23-0005.1

Mizukami, N., Rakovec, O., Newman, A. J., Clark, M. P., Wood, A. W., Gupta, H. V., & Kumar, R. (2019). On the choice of calibration metrics for "high-flow" estimation using hydrologic models. *Hydrology and Earth System Sciences*, *23*(6), 2601–2614. https://doi.org/10.5194/hess-23-2601-2019

Mohammadi, B. (2021). A review on the applications of machine learning for runoff modeling. *Sustainable Water Resources Management*, *7*(6), 98. https://doi.org/10.1007/s40899-021-00584-y

Nearing, G. S., Kratzert, F., Sampson, A. K., Pelissier, C. S., Klotz, D., Frame, J. M., Prieto, C., & Gupta, H. V. (2021). What Role Does Hydrological Science Play in the Age of Machine Learning? *Water Resources Research*, *57*(3), e2020WR028091. https://doi.org/10.1029/2020WR028091

Nepal, S., Flügel, W.-A., Krause, P., Fink, M., & Fischer, C. (2017). Assessment of spatial transferability of process-based hydrological model parameters in two neighbouring catchments in the Himalayan Region. *Hydrological Processes*, *31*(16), 2812–2826. https://doi.org/10.1002/hyp.11199

Parton, W. J., Holland, E. A., Del Grosso, S. J., Hartman, M. D., Martin, R. E., Mosier, A. R., Ojima, D. S., & Schimel, D. S. (2001). Generalized model for NO x and N2O emissions from soils. *Journal of Geophysical Research: Atmospheres*, *106*(D15), 17403–17419. https://doi.org/10.1029/2001JD900101

Porporato, A., Feng, X., Manzoni, S., Mau, Y., Parolari, A. J., & Vico, G. (2015). Ecohydrological modeling in agroecosystems: Examples and challenges. *Water Resources Research*, *51*(7), 5081–5099. https://doi.org/10.1002/2015WR017289

Raissi, M., Perdikaris, P., & Karniadakis, G. E. (2019). Physics-informed neural networks: A deep learning framework for solving forward and inverse problems involving nonlinear partial differential equations. *Journal of Computational Physics*, *378*, 686–707. https://doi.org/10.1016/j.jcp.2018.10.045

Read, J. S., Jia, X., Willard, J., Appling, A. P., Zwart, J. A., Oliver, S. K., Karpatne, A., Hansen, G. J. A., Hanson, P. C., Watkins, W., Steinbach, M., & Kumar, V. (2019). Process-Guided Deep Learning Predictions of Lake Water Temperature. *Water Resources Research*, *55*(11), 9173–9190. https://doi.org/10.1029/2019WR024922

Reichstein, M., Camps-Valls, G., Stevens, B., Jung, M., Denzler, J., Carvalhais, N., & Prabhat. (2019). Deep learning and process understanding for data-driven Earth system science. *Nature*, *566*(7743), 195–204. https://doi.org/10.1038/s41586-019-0912-1

Shen, C., Chen, X., & Laloy, E. (2021). Editorial: Broadening the Use of Machine Learning in Hydrology. *Frontiers in Water*, *3*. https://doi.org/10.3389/frwa.2021.681023

Sun, R., Pan, B., & Duan, Q. (2023). A surrogate modeling method for distributed land surface hydrological models based on deep learning. *Journal of Hydrology*, *624*, 129944. https://doi.org/10.1016/j.jhydrol.2023.129944

Wang, N., Zhang, D., Chang, H., & Li, H. (2020). Deep learning of subsurface flow via theory-guided neural network. *Journal of Hydrology*, *584*, 124700. https://doi.org/10.1016/j.jhydrol.2020.124700

Wang, Y.-H., & Gupta, H. V. (2024a). A Mass-Conserving-Perceptron for Machine-Learning-Based Modeling of Geoscientific Systems. *Water Resources Research*, *60*(4), e2023WR036461. https://doi.org/10.1029/2023WR036461

Wang, Y.-H., & Gupta, H. V. (2024b). Towards Interpretable Physical-Conceptual Catchment-Scale Hydrological Modeling Using the Mass-Conserving-Perceptron. *Water Resources Research*, *60*(10), e2024WR037224. https://doi.org/10.1029/2024WR037224

Wiggerthale, J., & Reich, C. (2024). Explainable Machine Learning in Critical Decision Systems: Ensuring Safe Application and Correctness. *AI*, *5*(4), Article 4. https://doi.org/10.3390/ai5040138

Wilson, A. G. (2025). *Deep Learning is Not So Mysterious or Different* (No. arXiv:2503.02113). arXiv. https://doi.org/10.48550/arXiv.2503.02113

Xiang, Z., Yan, J., & Demir, I. (2020). A Rainfall-Runoff Model With LSTM-Based Sequence-to-Sequence Learning. *Water Resources Research*, *56*(1), e2019WR025326. https://doi.org/10.1029/2019WR025326

Zhi, W., Appling, A. P., Golden, H. E., Podgorski, J., & Li, L. (2024). Deep learning for water quality. *Nature Water*, *2*(3), 228–241. https://doi.org/10.1038/s44221-024-00202-z

# Supplementary Information

## 1 Supplementary Methods

### 1.1 Study Area and Data Sources

This study applies the proposed framework to the **Samish subbasin**, located within the Puget Sound watershed in the northwestern United States.

**Elevation**
Elevation data (30m) were obtained from the USGS National Elevation Dataset (USGS 2019). Preprocessing, including dredging and flattening, was conducted using the JPDEM program (USEPA 2017).

**Land Cover and Permeability Fraction**
Land cover data were taken from the USGS National Land Cover Database (USGS 2024) at 90 m resolution. Land cover types were simplified for model representation. Permeability fractions for developed classes were assigned based on average perviousness values reported by USGS (2024). Forest composition was refined with the LEMMA dataset (2017), which provided alder distribution. Where alder was present, the alder class replaced the original USGS land cover classification.

**Forest Age and Harvest Disturbances**
Initial forest age maps were obtained from LEMMA (2017). Harvest disturbances were identified by comparing the 1990 and 2010 LEMMA age maps.

**Soil**
Soil properties were derived from two sources. Restrictive depth and soil organic carbon content were obtained from the NRCS SOLUS dataset (NRCS 2024).

**Meteorological Forcing**
Daily mean precipitation and temperature data were obtained from an 800 m resolution gridded dataset (Daly et al. 2021).

**Septic Systems**
Parcel-scale septic system data used in this study were not available to the public, while coarser census block–scale data are publicly accessible (Peterson et al. 2025). These data provide septic tank locations, and associated water and nitrate-as-ammonia loads were estimated assuming an average household size.

**Data for Model Calibration**

- **Hydrology**: Daily streamflow records (2009-2019) were obtained from the USGS hydrological monitoring station 12201500 (USGS 2021).

- **Water quality**: Monthly samples of ammonia, nitrate, dissolved organic carbon, and water temperature between 2009 and 2019 were provided by the Washington State Department of Ecology (2025).

**References for Data**

- Daly, C., Doggett, M., Smith, J., Olson, K., Halbleib, M., Dimcovic, Z., Keon, D., Loiselle, R., Steinberg, B., Ryan, A., Pancake, C., & Kaspar, E. (2021). Challenges in Observation-Based Mapping of Daily Precipitation across the Conterminous United States. *Journal of Atmospheric and Oceanic Technology*, 38(11), 1979–1992. https://doi.org/10.1175/JTECH-D-21-0054.1
- Peterson, B.K., Gordon, S.E., Williams, B.M., Atkins, R.M., Ahmed, L., & Seawolf, S.M. (2025). Estimated Densities of Residential Septic Tanks across the Conterminous United States for 12-digit Hydrologic Unit Code 12 (HUC12), National Hydrography Dataset Plus Version 2 (NHDPlusV2) Catchment, and Block Group Scales: U.S. Geological Survey data release. https://doi.org/10.5066/P1WCYDPB
- LEMMA (2017). GNN Structure (Species-Size) Maps. https://lemmadownload.forestry.oregonstate.edu/
- NRCS (2024). Soil Landscapes of the United States (SOLUS). https://www.nrcs.usda.gov/resources/data-and-reports/soil-landscapes-of-the-united-states-solus
- USEPA (2017). JPDEM User Manual. https://19january2021snapshot.epa.gov/water-research/velma-20-jpdem-user-manual-pdf_.html
- USGS (2019). National Elevation Dataset 30 Meter Tiles. https://gdg.sc.egov.usda.gov/Catalog/ProductDescription/NED.html
- USGS (2021). Samish River near Burlington, WA – 12201500. https://waterdata.usgs.gov/monitoring-location/12201500/
- USGS (2024). National Land Cover Database (NLCD) 2019 Products. https://www.usgs.gov/data/national-land-cover-database-nlcd-2019-products-ver-30-february-2024
- Washington State Department of Ecology. (2025). Water Quality Monitoring Station: 03B050 – Samish River near Burlington. https://apps.ecology.wa.gov/eim/search/SMP/RiverStreamSingleStationOverview.aspx?LocationUserIds=03B050&ResultType=RiverStreamOverviewList

## 1.2 Computational Environment

All models were executed (in Phase I and II) on:

- Intel® i9-13900 CPU @ 2.00 GHz
- 64 GB RAM
- NVIDIA RTX A4500 GPU (20 GB VRAM)

## 1.3 Model Characteristics in Phase I

Two representative process-based model types were analyzed: semi-distributed (SWAT) and fully distributed (VELMA). Their baseline computational characteristics are summarized in **Table S1**.

**Table S1. Computational characteristics of process-based models.**

| Model | Type | Spatial Resolution / Units | Key Outputs | Runtime |
|---|---|---|---|---|
| **SWAT** | Semi-distributed | 1,896 HRUs | Streamflow, nitrate | ~152 s |
| **VELMA** | Fully distributed | 244,933 grid cells (30 m) | Streamflow, nitrate | ~2 d 23 h |

To improve efficiency, resolution coarsening and process reduction were applied (**Table S2**).

**Table S2. Computational savings from model simplification.**

| Model | Simplification | Resolution / Units | Runtime | Notes |
|---|---|---|---|---|
| **HBV** | Process reduction | Single node | ~5 s | Lumped, Simplified N-cycle |
| **SWAT** | HRUs reduced | 187 HRUs | ~41 s | Reduced heterogeneity |
| **VELMA** | Grid downscaled | 360 m (1,748 cells) | ~15.2 min | Lower resolution, faster runtime |

Direct surrogate modeling, residual learning, transfer learning and hybrid strategies were applied to capture model behavior. Predictive performance was evaluated using the Nash–Sutcliffe efficiency (NSE) and Kling–Gupta efficiency (KGE). Computational efficiency was assessed by runtime, including both training and prediction (**Table S3**).

**Table S3. Training (2010-20114) and prediction (2015-2019) time of model–learning strategies.**

| Model Strategy | Training Period | Prediction Period | NSE-KGE-Flow | NSE-KGE-NO3 |
|---|---|---|---|---|
| **Direct (VELMA)** | 129.3 s | 5.2 s | 0.9575 | 0.5460 |
| **Direct (SWAT)** | 132.4 s | 5.5 s | 0.9255 | 0.8885 |
| **RC + TL (VELMA)** | 18.2 min | 7.7 min | 0.9640 | 0.9115 |
| **RC + TL (SWAT)** | 213.1 s | 29.5 s | 0.9345 | 0.9455 |
| **RC + RL (VELMA)** | 17.3 min | 15.2 min | 0.9860 | 0.8505 |
| **RC + RL (SWAT)** | 175.6 s | 51.7 s | 0.9950 | 0.9760 |
| **PR + RL (VELMA)** | 129.4 s | 12.6 s | 0.9165 | 0.5750 |
| **PR + RL (SWAT)** | 131.9 s | 13.2 s | 0.9645 | 0.9250 |
| **Hybrid (VELMA)** | 18.3 min | 13.5 s | 0.9405 | 0.8440 |
| **Hybrid (SWAT)** | 217.8 s | 15.6 s | 0.9690 | 0.9515 |

**Notes:** RC = Resolution Coarsening; PR = Process Reduction; TL = Transfer Learning; RL = Reinforcement Learning; Hybrid = Combination of RC/PR with TL or RL; NSE-KGE = average Nash–Sutcliffe Efficiency and Kling–Gupta Efficiency metrics for flow and nitrate ($NO_3^-$) Loss predictions.

## 1.4 Simplified Nitrogen Module added to HBV

We developed a lumped, daily time-step nitrogen (N) module coupled to simulated streamflow. The module represents three conceptual pools—ammonium ($NH_4^+$), nitrate ($NO_3^-$), and organic nitrogen (OrgN)—and accounts for external inputs, plant uptake, mineralization, temperature-dependent losses, and hydrologic export. A first-order reservoir simulates lagged $NO_3^-$ signals at the outlet. Detailed descriptions and governing equations are provided in *Nitrogen_Module_HBV.md (locate in* https://github.com/jlonghku/EcoHydroModel/doc*)*.

## 1.5 Nitrification equations in Phase II

We implemented two process-based nitrification modules grounded in established biogeochemical studies. The first follows **Del Grosso–type formulations**, where daily nitrification fluxes are computed as functions of ammonium availability, humus decomposition, and environmental modifiers (water-filled pore space, temperature, and pH). Soil-specific coefficients capture heterogeneity across soil types. The second follows **Parton–type formulations**, linking nitrification fluxes to ammonium availability and humus decomposition, further modulated by soil moisture, temperature, and pH. Here, soil- and layer-specific parameters explicitly represent heterogeneity in depth, porosity, and bulk density, with unit conversions applied between mass- and area-based forms. Detailed formulations and governing equations are provided at https://github.com/jlonghku/EcoHydroModel/doc.

- *Nitrification_Module_Del_Grosso.md*
- *Nitrification_Module_Parton.md*

## 1.6 Deep Learning Model architecture

In Phase I, we developed a unified deep learning model for residual and transfer learning. During transfer learning, all layers except the output heads were frozen to preserve learned representations while adapting to new data. The architecture, implemented in PyTorch, comprised:

- Two LSTM layers (256 units each) with LayerNorm for temporal feature encoding.
- Shared fully connected layers (256 and 128 units) for feature extraction.
- Dual output heads:
    - **Streamflow**: a fully connected linear output layer.
    - **Nitrate loss**: four stacked LSTM layers followed by fully connected branches. The model was trained for 100 epochs using the Adam optimizer with a learning rate of 0.001.

In Phase II, we employed an alternative deep learning model based on a multilayer perceptron (MLP). The architecture included:

- An embedding layer for soil type, concatenated with dynamic state features.
- A LayerNorm layer applied to the combined inputs.
- A feedforward network with four fully connected hidden layers (64 units each, ReLU activation).
- A fully connected linear output layer producing nitrification rates. This model was also trained for 300 epochs with the Adam optimizer at a learning rate of 0.001.

# 2 Supplementary Code and Documentation

We provide a Python package implementing the **EcoHydroModel** modeling framework: GitHub repository (https://github.com/jlonghku/EcoHydroModel).